\documentclass{article}

\usepackage[final]{neurips_2024_calm}
\usepackage{subcaption}

\usepackage[utf8]{inputenc} % allow utf-8 input
\usepackage[T1]{fontenc}    % use 8-bit T1 fonts
\usepackage{hyperref}       % hyperlinks
\usepackage{url}            % simple URL typesetting
\usepackage{tabu} 
\usepackage{booktabs}       % professional-quality tables
\usepackage{amsfonts}       % blackboard math symbols
\usepackage{nicefrac}       % compact symbols for 1/2, etc.
\usepackage{microtype}      % microtypography
\usepackage{xcolor}         % colors
\usepackage{graphicx}
\usepackage{caption}
\usepackage{subcaption}
\usepackage{float}
\usepackage{multirow}
\usepackage{mathtools}
\usepackage{paralist, tabularx}
\usepackage{float}
\usepackage{placeins}
\usepackage{comment}
\usepackage{svg}
\usepackage{xcolor}

\title{Causal Interventions on Causal Paths: Mapping GPT-2's Reasoning From Syntax to Semantics}

\newcommand{\aspace}{\hspace{2em}}

\author{%
  \textbf{Isabelle Lee} \aspace
  \textbf{Joshua Lum} \aspace
  \textbf{Ziyi Liu} \aspace
  \textbf{Dani Yogatama}\\
  \\
  \vspace{1mm}
  University of Southern California\\
  \vspace{1mm}
  \texttt{lee.isabelle.g@gmail.com} 
}

\begin{document}
\maketitle
\begin{abstract}

While interpretability research has shed light on some internal algorithms utilized by transformer-based LLMs, reasoning in natural language, with its deep contextuality and ambiguity, defies easy categorization. As a result, formulating clear and motivating questions for circuit analysis that rely on well-defined in-domain and out-of-domain examples required for causal interventions is challenging. Although significant work has investigated circuits for specific tasks, such as indirect object identification (IOI), deciphering natural language reasoning through circuits remains difficult due to its inherent complexity. In this work, we take initial steps to characterize causal reasoning in LLMs by analyzing clear-cut cause-and-effect sentences like "I opened an \textcolor{blue}{umbrella} \textcolor{teal}{because} it started \textcolor{purple}{raining}," where causal interventions may be possible through carefully crafted scenarios using GPT-2 small. Our findings indicate that causal syntax is localized within the first 2-3 layers, while certain heads in later layers exhibit heightened sensitivity to nonsensical variations of causal sentences. This suggests that models may infer reasoning by (1) detecting syntactic cues and (2) isolating distinct heads in the final layers that focus on semantic relationships.

\end{abstract}

\section{Introduction}
As transformer-based large language models (LLMs) scale up, their performance on diverse downstream tasks has shown remarkable improvement \citep{Wei2022EmergentAO, Srivastava2022BeyondTI}. 
These models demonstrate remarkable capabilities across various tasks, from reasoning tasks such as math problem solving and commonsense reasoning to question-answering that require knowledge synthesis \cite{Kojima2022LargeLM, Zellers2018FromRT,Wei2022ChainOT,Brown2020LanguageMA}. Understanding and benchmarking these capabilities has become a prolific research area, as both technical communities and the general public uncover new ways to harness LLMs. 
Despite these impressive abilities, however, the mechanisms driving these capabilities remain largely opaque.

As model scales increase, interpreting their associated capabilities becomes increasingly challenging. 
Nevertheless, notable advancements in interpretability have improved our understanding of these models. 
Recent work in mechanistic interpretability takes a microscopic approach to analyze models \citep{olah2020zoom}. 
Many studies derive interpretable features and behaviors from attention mechanisms using simplified toy models of transformers, revealing concepts like induction heads and in-context learning \citep{olsson2022context, elhage2021mathematical}. 
Although these insights shed light on interpretable, microscopic mechanisms like feature recognition and copying, they fall short in explaining complex, high-level behaviors in realistic tasks.
A major reason for this is that circuits rely on causal interventions, which require clear distinctions between in-domain and out-of-domain examples. 
However, many natural language tasks are complex and inherently ambiguous; for instance, the distinction between reasoning and non-reasoning text is often murky. 
This ambiguity complicates the scaling of interpretability efforts to such macroscopic behaviors. 
%Additional challenges include deciphering the learned representations at different model layers and understanding how these representations influence macroscopic behavioral outcomes.

To begin to understand macroscopic reasoning behaviors of LLMs and link them to underlying representations, we focus on the simplest cases of reasoning by breaking them down into components like cause-and-effect relations. 
Specifically, we examine clear-cut causal phrases connected by markers such as "because" and "so." 
We design scenarios that allow for causal interventions and investigate whether model responses—such as patterns observed in attention maps and logit shifts in the residual stream—can be traced to these semantic perturbations.

\begin{figure}[h]
\centering
\begin{subfigure}[b]{1\textwidth}
   \includegraphics[width=1\textwidth]{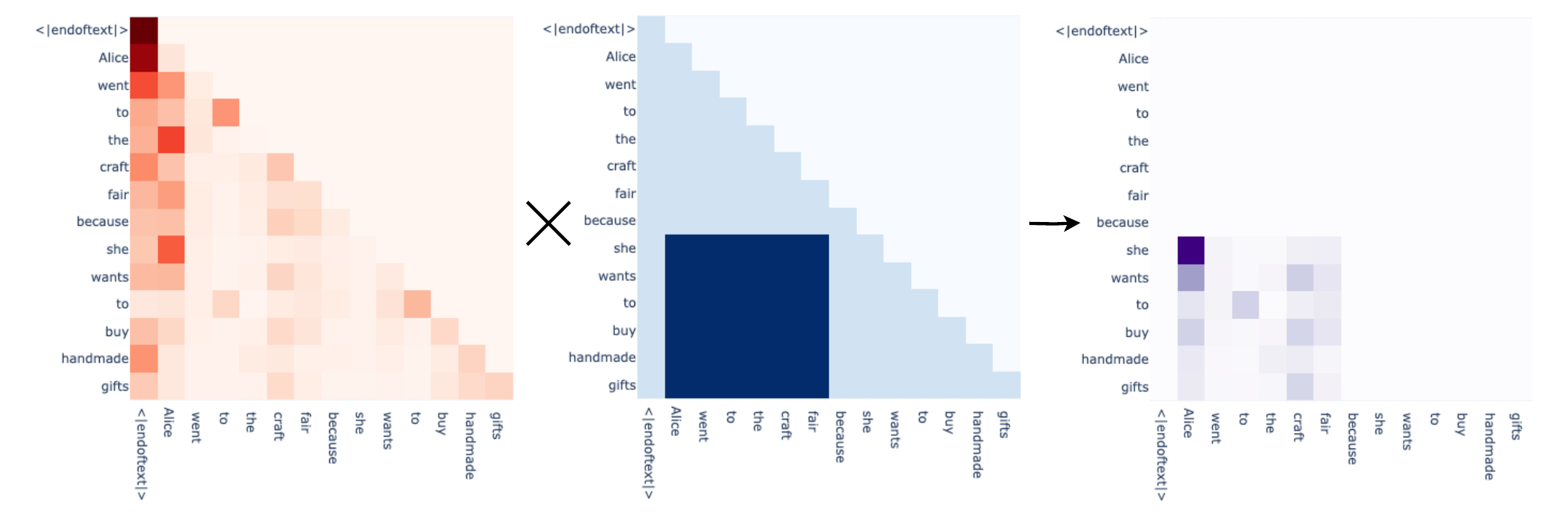}
   \caption{Attention Analysis}
   \label{fig:attn-analysis} 
\end{subfigure}
\begin{subfigure}[b]{0.7\textwidth}
   \includegraphics[width=1\linewidth]{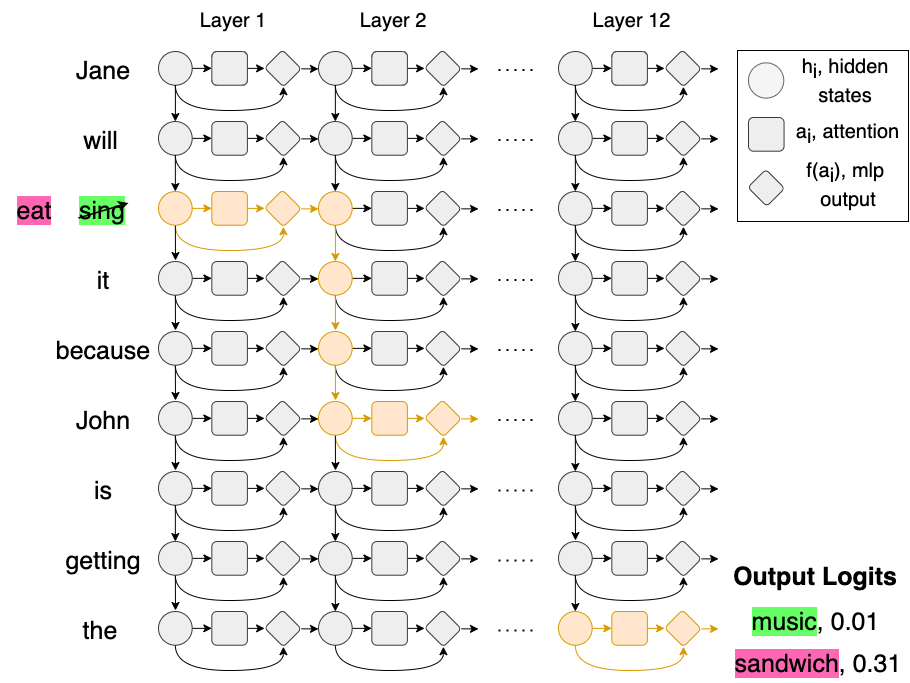}
   \caption{Activation patching with an example causal trace highlighted in orange}
   \label{fig:act-patch} 
\end{subfigure}
\caption{Overview of methods.}
\label{fig:overview}
\end{figure}

\paragraph{In this work,} 
We analyze GPT-2’s ability to comprehend causal relationships in sentences with clear, unambiguous causal connections. In these cases, we anticipate that introducing nonsensical perturbations will reveal distinct causal circuits within the model. Our focus is on straightforward instances where action verbs interact causally with specific settings (e.g., locations) or plausible objects. We find that GPT-2 primarily captures syntactic structures within its first 2-3 layers. We then perform causal interventions on the model's semantic activations to identify which attention heads contribute to task performance. Our results reveal that a small set of attention heads consistently activates across subtasks. Future work could explore more complex causal scenarios or sentences with ambiguous causal relationships and compare these findings with larger models to determine if similar patterns emerge across different settings.

\section{Overview}
We explore how LLMs understand reasoning by examining their responses to sentences with straightforward reasoning structures. 
We conduct our experiments with GPT-2 small, a 12-layer model with decoder blocks containing self-attention layers with 12 attention heads and multilayer perceptrons (MLPs) \citep{Radford2019LanguageMA}. 
We recognize that humans comprehend reasoning in natural language in two steps. 
First, by identifying syntactic cues associated with reasoning, such as phrases connected by words like "because" and "so", we assess whether a sentence likely contains reasoning relations. 
Next, we consider the semantic relationships within cause-and-effect phrases. 
Our experiments are designed to reflect this two-step reasoning process. 
For syntactic analysis, we use a dataset of diverse sentence structures (see Table \ref{tab:syntax-data}). 
For semantic analysis, we modify cause-and-effect phrases in templated sentences (see Table \ref{tab:act-patch-data}) to make the reasoning relations either coherent or nonsensical.

\section{Where Is Syntax in a Transformer?}
To locate syntactical knowledge in GPT-2, we analyze the model responses to a curated synthetic dataset of causal sentences with varying syntax. We generated the dataset by prompting the language models with multiple templates, as summarized in Table \ref{tab:syntax-data}. We assess attention patterns based on the causal phrases and delimiters, following an approach similar to the syntactical analysis performed by \cite{Vig2019AnalyzingTS} on BERT.

\paragraph{Setup and Methods}

The templates used to generate the syntactical dataset in Table \ref{tab:syntax-data} show the syntactical structure of the sentences in the form of $[ \mathcolor{blue}{e_1, \cdots, e_n}, \mathcolor{teal}{d}, \mathcolor{purple}{c_1, \cdots c_m}]$ or $[ \mathcolor{purple}{c_1, \cdots c_m}, \mathcolor{teal}{d}, \mathcolor{blue}{e_1, \cdots, e_n}]$ where $c_i = $ tokens of a cause phrase, $d = $ causal delimiter token, and $e_j = $ tokens of an effect phrase. 
Respectively, the first template refers to "because" sentences and the second template refers to "so" sentences.
An example of such causal sentence is ``\textcolor{blue}{Alice went to the craft fair} \textcolor{teal}{because} \textcolor{purple}{she wants to buy handmade gifts}." Then, we specifically analyze the attention maps by calculating 1) how much attention is paid to the causal delimiters and 2) how much effect token attends to cause tokens. We calculate 1) as 
\begin{equation}
    P_d = \frac{ \sum_{j=1}^{m} \alpha_{d,j}}{\sum_{i=1}^{n+m+1} \sum_{j=1}^{n+m+1} \alpha_{i,j}},
\end{equation}
where $\alpha_{i,j} = \left[ \text{softmax}\left( QK^T/\sqrt{d_K} \right)V \right]_{i,j}$ with query $Q$, key $K$, and value $V$ matrices calculated from the input tokens with attention weights with $1/\sqrt{d_K}$ as a scaling factor calculated from the dimension of the key matrix.
We then calculate 2) proportion of cause-to-effect or effect-to-cause attention similarly. As described in Figure \ref{fig:attn-analysis}, we isolate the cause-to-effect or effect-to-cause attention patterns by masking. 
The proportion of causal attention pattern can be expressed as 
\begin{equation}
    P_{c} = \frac{ \sum_{i=1}^{n} \sum_{j=1}^{m} \alpha_{i,j}}{\sum_{i=1}^{n+m+1} \sum_{j=1}^{n+m+1} \alpha_{i,j}}.
\end{equation}
With isolated causal attention map, we perform statistical analyses per head per layers.

\subsection{Results}
\begin{figure}[h]
\centering
\begin{subfigure}[b]{0.3\textwidth}
   \includegraphics[width=1\linewidth]{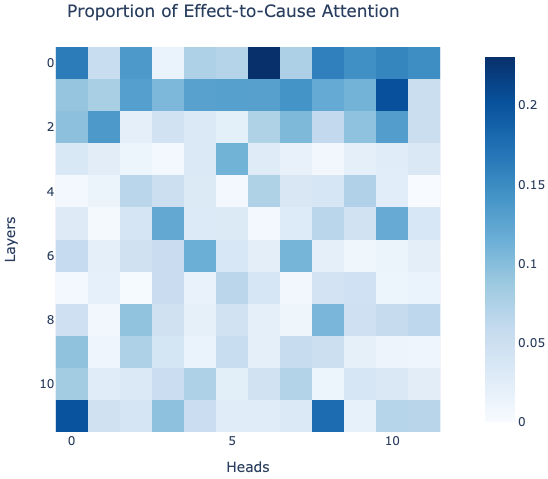}
   \caption{Because}
   \label{fig:e2c-because} 
\end{subfigure}
\begin{subfigure}[b]{0.3\textwidth}
   \includegraphics[width=1\linewidth]{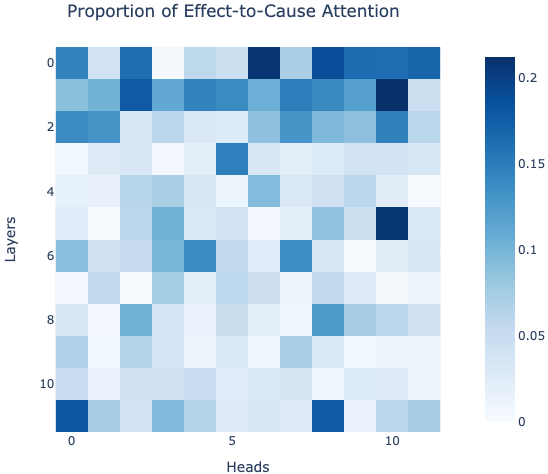}
   \caption{So}
   \label{fig:e2c-so} 
\end{subfigure}
\begin{subfigure}[b]{0.3\textwidth}
   \includegraphics[width=1\linewidth]{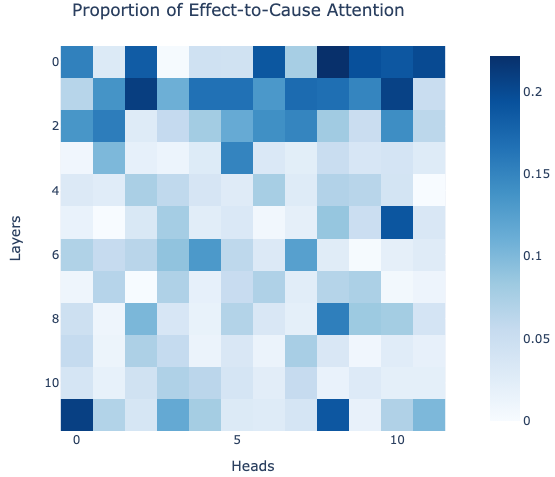}
   \caption{Therefore}
   \label{fig:e2c-therefore} 
\end{subfigure}
\begin{subfigure}[b]{0.3\textwidth}
   \includegraphics[width=1\linewidth]{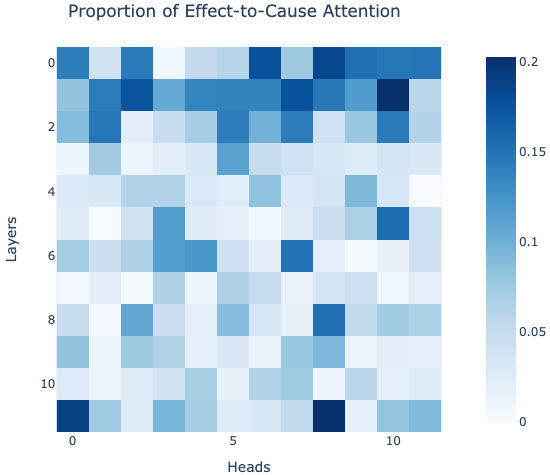}
   \caption{Resulting}
   \label{fig:e2c-resulting} 
\end{subfigure}
\begin{subfigure}[b]{0.3\textwidth}
   \includegraphics[width=1\linewidth]{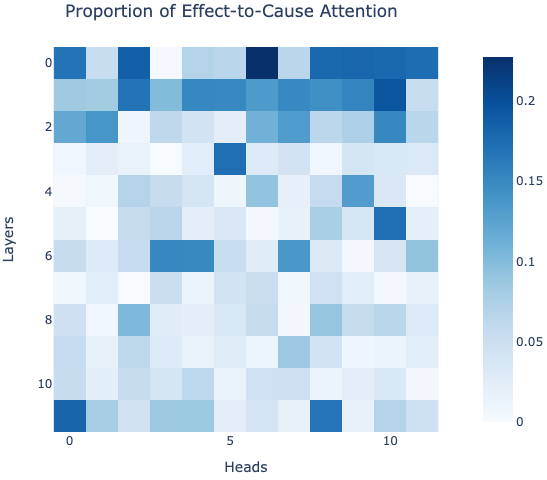}
   \caption{Since}
   \label{fig:e2c-since} 
\end{subfigure}
\caption{Proportion of Effect-to-Cause or Cause-to-Effect Attention.}
\label{fig:attn-e2c}
\end{figure}

In order to analyze syntactical understanding of GPT-2, we first compute the proportion of attention paid to causal delimiters, $P_d$, such as ``because" and ``so." Figure \ref{fig:attn-syntax} summarizes the results, which shows that the heads that pay attention to delimiters are spread across the layers with some concentrations in the earlier layers of a transformer. On the other hand, Figure \ref{fig:attn-e2c} shows that the heads that pay particular causal attention, $P_c$, tend to be most concentrated in the first 2-3 layers.

\begin{comment}
\subsection{Some Observations of the Attention Maps from the First 3 Layers}
\end{comment}

%\subsection{"Because" and "So" Neurons}

\section{Locating Semantics: Where Does GPT-2 Figure Out a Sandwich Is for Eating, not Singing?}
We also consider logit analysis at each layer of the model to analyze model behavior with causal sentences. From the residual stream, we calculate the per token loss at each layer, which we define as per layer loss. 
We can then hypothesize that the causal relations between two phrases in a sentence would be reflected in the per-layer logit calculation. 
An illustrated example of per layer loss calculation is shown in Figure \ref{fig:pll}.
In this example, we take a causal sentence ``we went shopping because we were bored" and perturb it to make a non-causal sentence. 
We swap ``bored" with ``sleepy" to obscure the causal relations between the two phrases of the sentence. 
In this work, we focus on scenarios where semantic perturbations can occur through straightforward word substitutions. 
Specifically, we examine sentences that involve an action verb in relation to a specific location or an action verb acting on a particular object. 
Because these relationships are causally specific and syntactically simple, we can easily distort the sentences to render them nonsensical, such as by replacing a location or object.
Our dataset is detailed in Table \ref{tab:act-patch-data}.
We see that perturbing a sentence this way is reflected in the per layer loss calculation.
With this overall analysis in mind, we can then decompose the residual contribution per attention heads, per neurons, and analyze their implications for finding causal relations.

\subsection{Activation Patching Results}

We apply activation patching to contrastive pairs of causal sentences. As outlined in \ref{fig:act-patch}, we first run our model using an original causal sentence. Next, we introduce a semantic perturbation by replacing the sentence with its contrastive pair and rerun the model. By tracking the activation differences that result in changes to the final logit predictions, we pinpoint specific model components responsible for distinguishing causal semantics from random semantics.

\begin{figure}[h]
    \centering
    \begin{subfigure}{0.32\textwidth}
        \centering
        \includegraphics[height=0.2\textheight]{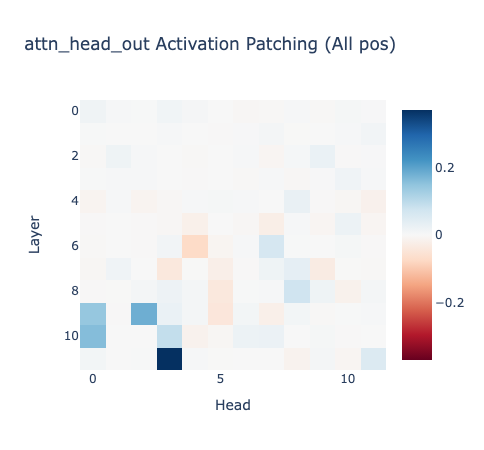}
        \caption{ALS template}
        \label{fig:sub1}
    \end{subfigure}
    \begin{subfigure}{0.32\textwidth}
        \centering
        \includegraphics[height=0.2\textheight]{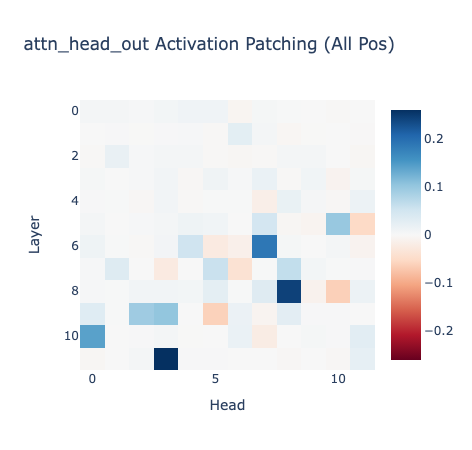}
        \caption{ALB template}
        \label{fig:sub2}
    \end{subfigure}
    \begin{subfigure}{0.32\textwidth}
        \centering
        \includegraphics[height=0.2\textheight]{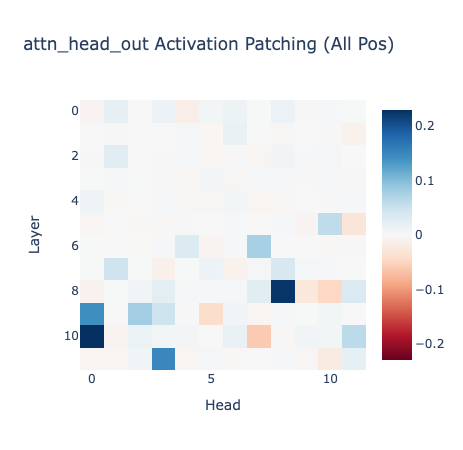}
        \caption{AOS template}
        \label{fig:sub2}
    \end{subfigure}
    \begin{subfigure}{0.32\textwidth}
        \centering
        \includegraphics[height=0.2\textheight]{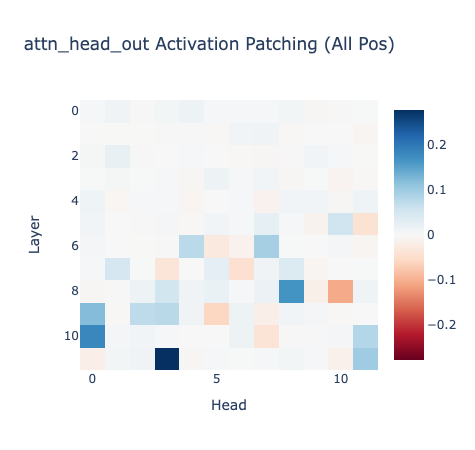}
        \caption{AOB template}
        \label{fig:sub2}
    \end{subfigure}
    \begin{subfigure}{0.32\textwidth}
        \centering
        \includegraphics[height=0.2\textheight]{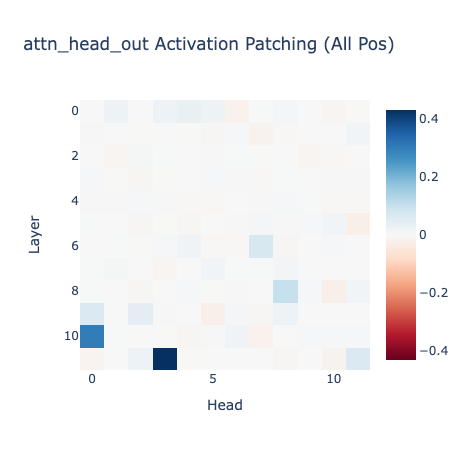}
        \caption{AOB template}
        \label{fig:sub2}
    \end{subfigure}
    \caption{Attention head out activation patching results over all positions. O = Object, L = location, S = So, B = Because}
    \label{fig:attn_head_activation}
\end{figure}

As shown in Figure \ref{fig:attn_head_activation}, few distinct attention heads in the middle to last few layers contribute most to the logit difference, especially layer 11 head 2, layer 10 head 0, and layer 8 head 8, light up in most templates. 
We also note that in the residual stream, the ``PERTUBRED'' token significantly influences predictions in the earlier layers, as shown in Figures \ref{fig:als_template}, \ref{fig:alb_template}, \ref{fig:als_with_template}, \ref{fig:aos_template}, and \ref{fig:aob_template}. 
%In the final two layers, the last token plays the most critical role in determining the next token. 
%Finally, it is consistently layer 0 in the MLP output that is activated by the perturbed token.

%\section{Causal Intervention for Causal Mechanisms}
%\input{sections/5-causal-intervention}

\section{Conclusion}

Our investigation suggests that the model demonstrates a syntactic focus in its initial layers, with attention mechanisms primarily engaging at this stage. As processing deepens, a shift occurs, and the model begins to handle reasoning tasks in a more semantic manner, particularly in the later layers. These findings are evident in cases of clear-cut reasoning, where causal relationships can be perturbed with word substitutions. However, ambiguity in reasoning presents a more complex challenge. Future work will aim to explore how the model adapts when faced with ambiguous or less structured reasoning tasks, as understanding these scenarios could significantly enhance the clarity of causal inference and model interpretability.

\section{Related Work}

\paragraph{Reasoning in LLMs} LLMs have demonstrated remarkable ``emergent" abilities for which they were not explicitly trained, though mechanisms behind them are not well understood \citep{Wei2022EmergentAO, Schaeffer2023AreEA,Lu2023AreEA}. Among them are LLMs' ability to reason in many domains from informal, commonsense reasoning \citep{Kojima2022LargeLM, Bhagavatula2019AbductiveCR,Zellers2018FromRT} to more formal domains such as scientific reasoning \citep{Lu2022LearnTE, Birhane2023ScienceIT} and mathematical reasoning \citep{Cobbe2021TrainingVT, Yuan2023ScalingRO}. Behavioral studies have focused significant recent efforts in characterizing and benchmarking model capabilities \citep{Srivastava2022BeyondTI, Huang2023CanLL}, but they are not well connected to the intermediate representations and internal responses of a model. Our work provides first steps in connecting behavioral observations to internal and mechanical model responses with curated tasks. %\cite{Wu2023InterpretabilityAS}

\paragraph{Attention Analysis and Mechanistic Interpretability} 
Attention maps have been used for interpreting intermediate representations and behaviors of transformers since the transformer architectures took off in language modeling \citep{Jain2019AttentionIN, Wiegreffe2019AttentionIN, Clark2019WhatDB, Rogers2020API}. Analyzing what the language models pay attention to when making predictions can elucidate relevant features for particular labels. 
Recently, work in mechanistic interpretability largely approximated transformers with simplified attention matrix multiplications to describe ``circuits" \citep{elhage2021mathematical}.
``Circuits" in LLMs can be thought of as information flow through a transformer that make certain decisions or perform a particular task.
%We use similar formulation of the transformer as described by \citet{elhage2021mathematical}, and analyze attention maps to locate ``syntax heads" which pays particular attention to syntactical delimiters of causal sentences.

\paragraph{Causal Tracing (Activation Patching) and Causal Intervention} While many recent behavioral characterizations of LLMs rely on post-hoc benchmarking, some interpretability methods actively engage with model responses. 
For instance, counterfactual perturbations on input data have been used to study subject-verb agreements in BERT by tracing model responses to particular input representations \citep{Ravfogel2021CounterfactualIR,Elazar2022MeasuringCE}.
First introduced by \cite{meng2022locating}, activation patching \textit{causally traces} the effect of perturbed input token on the activations throughout the layers and eventually on the predicted output token.
Activation patching has been used to locate factual information in a transformer in the case of \cite{meng2022locating}, and it is frequently used for identifying circuits in LLMs.
\cite{wang2022interpretability} used activation patching to identify a circuit that performs the ``indirect object identification task," in which a model predicts the name as object of an action given the previous context.
%More concretely, the task is to complete sentences like ``When Mary and John went to the store, John gave a drink to" with ``Mary" rather than ``John." 
%With particular model components involved in a particular circuit identified, we may perform causal interventions such as ablations to measure their direct or indirect effect on the logits or task performance.

\bibliography{neurips}

\begin{thebibliography}{27}
\providecommand{\natexlab}[1]{#1}
\providecommand{\url}[1]{\texttt{#1}}
\expandafter\ifx\csname urlstyle\endcsname\relax
  \providecommand{\doi}[1]{doi: #1}\else
  \providecommand{\doi}{doi: \begingroup \urlstyle{rm}\Url}\fi

\bibitem[Bhagavatula et~al.(2019)Bhagavatula, Bras, Malaviya, Sakaguchi, Holtzman, Rashkin, Downey, Yih, and Choi]{Bhagavatula2019AbductiveCR}
Chandra Bhagavatula, Ronan~Le Bras, Chaitanya Malaviya, Keisuke Sakaguchi, Ari Holtzman, Hannah Rashkin, Doug Downey, Scott Yih, and Yejin Choi.
\newblock Abductive commonsense reasoning.
\newblock \emph{ArXiv}, abs/1908.05739, 2019.
\newblock URL \url{https://api.semanticscholar.org/CorpusID:201058651}.

\bibitem[Birhane et~al.(2023)Birhane, Kasirzadeh, Leslie, and Wachter]{Birhane2023ScienceIT}
Abeba Birhane, Atoosa Kasirzadeh, David Leslie, and Sandra Wachter.
\newblock Science in the age of large language models.
\newblock \emph{Nature Reviews Physics}, 5:\penalty0 277--280, 2023.
\newblock URL \url{https://api.semanticscholar.org/CorpusID:258361324}.

\bibitem[Brown et~al.(2020)Brown, Mann, Ryder, Subbiah, Kaplan, Dhariwal, Neelakantan, Shyam, Sastry, Askell, Agarwal, Herbert-Voss, Krueger, Henighan, Child, Ramesh, Ziegler, Wu, Winter, Hesse, Chen, Sigler, Litwin, Gray, Chess, Clark, Berner, McCandlish, Radford, Sutskever, and Amodei]{Brown2020LanguageMA}
Tom~B. Brown, Benjamin Mann, Nick Ryder, Melanie Subbiah, Jared Kaplan, Prafulla Dhariwal, Arvind Neelakantan, Pranav Shyam, Girish Sastry, Amanda Askell, Sandhini Agarwal, Ariel Herbert-Voss, Gretchen Krueger, Tom Henighan, Rewon Child, Aditya Ramesh, Daniel~M. Ziegler, Jeff Wu, Clemens Winter, Christopher Hesse, Mark Chen, Eric Sigler, Mateusz Litwin, Scott Gray, Benjamin Chess, Jack Clark, Christopher Berner, Sam McCandlish, Alec Radford, Ilya Sutskever, and Dario Amodei.
\newblock Language models are few-shot learners.
\newblock \emph{ArXiv}, abs/2005.14165, 2020.
\newblock URL \url{https://api.semanticscholar.org/CorpusID:218971783}.

\bibitem[Clark et~al.(2019)Clark, Khandelwal, Levy, and Manning]{Clark2019WhatDB}
Kevin Clark, Urvashi Khandelwal, Omer Levy, and Christopher~D. Manning.
\newblock What does bert look at? an analysis of bert’s attention.
\newblock In \emph{BlackboxNLP@ACL}, 2019.
\newblock URL \url{https://api.semanticscholar.org/CorpusID:184486746}.

\bibitem[Cobbe et~al.(2021)Cobbe, Kosaraju, Bavarian, Chen, Jun, Kaiser, Plappert, Tworek, Hilton, Nakano, Hesse, and Schulman]{Cobbe2021TrainingVT}
Karl Cobbe, Vineet Kosaraju, Mohammad Bavarian, Mark Chen, Heewoo Jun, Lukasz Kaiser, Matthias Plappert, Jerry Tworek, Jacob Hilton, Reiichiro Nakano, Christopher Hesse, and John Schulman.
\newblock Training verifiers to solve math word problems.
\newblock \emph{ArXiv}, abs/2110.14168, 2021.
\newblock URL \url{https://api.semanticscholar.org/CorpusID:239998651}.

\bibitem[Elazar et~al.(2022)Elazar, Kassner, Ravfogel, Feder, Ravichander, Mosbach, Belinkov, Schutze, and Goldberg]{Elazar2022MeasuringCE}
Yanai Elazar, Nora Kassner, Shauli Ravfogel, Amir Feder, Abhilasha Ravichander, Marius Mosbach, Yonatan Belinkov, Hinrich Schutze, and Yoav Goldberg.
\newblock Measuring causal effects of data statistics on language model's 'factual' predictions.
\newblock \emph{ArXiv}, abs/2207.14251, 2022.
\newblock URL \url{https://api.semanticscholar.org/CorpusID:251134985}.

\bibitem[Elhage et~al.(2021)Elhage, Nanda, Olsson, Henighan, Joseph, Mann, Askell, Bai, Chen, Conerly, DasSarma, Drain, Ganguli, Hatfield-Dodds, Hernandez, Jones, Kernion, Lovitt, Ndousse, Amodei, Brown, Clark, Kaplan, McCandlish, and Olah]{elhage2021mathematical}
Nelson Elhage, Neel Nanda, Catherine Olsson, Tom Henighan, Nicholas Joseph, Ben Mann, Amanda Askell, Yuntao Bai, Anna Chen, Tom Conerly, Nova DasSarma, Dawn Drain, Deep Ganguli, Zac Hatfield-Dodds, Danny Hernandez, Andy Jones, Jackson Kernion, Liane Lovitt, Kamal Ndousse, Dario Amodei, Tom Brown, Jack Clark, Jared Kaplan, Sam McCandlish, and Chris Olah.
\newblock A mathematical framework for transformer circuits.
\newblock \emph{Transformer Circuits Thread}, 2021.
\newblock https://transformer-circuits.pub/2021/framework/index.html.

\bibitem[Huang et~al.(2023)Huang, Mamidanna, Jangam, Zhou, and Gilpin]{Huang2023CanLL}
Shiyuan Huang, Siddarth Mamidanna, Shreedhar Jangam, Yilun Zhou, and Leilani Gilpin.
\newblock Can large language models explain themselves? a study of llm-generated self-explanations.
\newblock \emph{ArXiv}, abs/2310.11207, 2023.
\newblock URL \url{https://api.semanticscholar.org/CorpusID:264172366}.

\bibitem[Jain and Wallace(2019)]{Jain2019AttentionIN}
Sarthak Jain and Byron~C. Wallace.
\newblock Attention is not explanation.
\newblock In \emph{North American Chapter of the Association for Computational Linguistics}, 2019.
\newblock URL \url{https://api.semanticscholar.org/CorpusID:67855860}.

\bibitem[Kojima et~al.(2022)Kojima, Gu, Reid, Matsuo, and Iwasawa]{Kojima2022LargeLM}
Takeshi Kojima, Shixiang~Shane Gu, Machel Reid, Yutaka Matsuo, and Yusuke Iwasawa.
\newblock Large language models are zero-shot reasoners.
\newblock \emph{ArXiv}, abs/2205.11916, 2022.
\newblock URL \url{https://api.semanticscholar.org/CorpusID:249017743}.

\bibitem[Lu et~al.(2022)Lu, Mishra, Xia, Qiu, Chang, Zhu, Tafjord, Clark, and Kalyan]{Lu2022LearnTE}
Pan Lu, Swaroop Mishra, Tony Xia, Liang Qiu, Kai-Wei Chang, Song-Chun Zhu, Oyvind Tafjord, Peter Clark, and A.~Kalyan.
\newblock Learn to explain: Multimodal reasoning via thought chains for science question answering.
\newblock \emph{ArXiv}, abs/2209.09513, 2022.
\newblock URL \url{https://api.semanticscholar.org/CorpusID:252383606}.

\bibitem[Lu et~al.(2023)Lu, Bigoulaeva, Sachdeva, Madabushi, and Gurevych]{Lu2023AreEA}
Sheng Lu, Irina Bigoulaeva, Rachneet Sachdeva, Harish~Tayyar Madabushi, and Iryna Gurevych.
\newblock Are emergent abilities in large language models just in-context learning?
\newblock \emph{ArXiv}, abs/2309.01809, 2023.
\newblock URL \url{https://api.semanticscholar.org/CorpusID:261531236}.

\bibitem[Meng et~al.(2022)Meng, Bau, Andonian, and Belinkov]{meng2022locating}
Kevin Meng, David Bau, Alex Andonian, and Yonatan Belinkov.
\newblock Locating and editing factual associations in gpt.
\newblock \emph{Advances in Neural Information Processing Systems}, 35:\penalty0 17359--17372, 2022.

\bibitem[Olah et~al.(2020)Olah, Cammarata, Schubert, Goh, Petrov, and Carter]{olah2020zoom}
Chris Olah, Nick Cammarata, Ludwig Schubert, Gabriel Goh, Michael Petrov, and Shan Carter.
\newblock Zoom in: An introduction to circuits.
\newblock \emph{Distill}, 2020.
\newblock \doi{10.23915/distill.00024.001}.
\newblock https://distill.pub/2020/circuits/zoom-in.

\bibitem[Olsson et~al.(2022)Olsson, Elhage, Nanda, Joseph, DasSarma, Henighan, Mann, Askell, Bai, Chen, Conerly, Drain, Ganguli, Hatfield-Dodds, Hernandez, Johnston, Jones, Kernion, Lovitt, Ndousse, Amodei, Brown, Clark, Kaplan, McCandlish, and Olah]{olsson2022context}
Catherine Olsson, Nelson Elhage, Neel Nanda, Nicholas Joseph, Nova DasSarma, Tom Henighan, Ben Mann, Amanda Askell, Yuntao Bai, Anna Chen, Tom Conerly, Dawn Drain, Deep Ganguli, Zac Hatfield-Dodds, Danny Hernandez, Scott Johnston, Andy Jones, Jackson Kernion, Liane Lovitt, Kamal Ndousse, Dario Amodei, Tom Brown, Jack Clark, Jared Kaplan, Sam McCandlish, and Chris Olah.
\newblock In-context learning and induction heads.
\newblock \emph{Transformer Circuits Thread}, 2022.
\newblock https://transformer-circuits.pub/2022/in-context-learning-and-induction-heads/index.html.

\bibitem[Radford et~al.(2019)Radford, Wu, Child, Luan, Amodei, and Sutskever]{Radford2019LanguageMA}
Alec Radford, Jeff Wu, Rewon Child, David Luan, Dario Amodei, and Ilya Sutskever.
\newblock Language models are unsupervised multitask learners.
\newblock 2019.
\newblock URL \url{https://api.semanticscholar.org/CorpusID:160025533}.

\bibitem[Ravfogel et~al.(2021)Ravfogel, Prasad, Linzen, and Goldberg]{Ravfogel2021CounterfactualIR}
Shauli Ravfogel, Grusha Prasad, Tal Linzen, and Yoav Goldberg.
\newblock Counterfactual interventions reveal the causal effect of relative clause representations on agreement prediction.
\newblock In \emph{Conference on Computational Natural Language Learning}, 2021.
\newblock URL \url{https://api.semanticscholar.org/CorpusID:234681155}.

\bibitem[Rogers et~al.(2020)Rogers, Kovaleva, and Rumshisky]{Rogers2020API}
Anna Rogers, Olga Kovaleva, and Anna Rumshisky.
\newblock A primer in bertology: What we know about how bert works.
\newblock \emph{Transactions of the Association for Computational Linguistics}, 8:\penalty0 842--866, 2020.
\newblock URL \url{https://api.semanticscholar.org/CorpusID:211532403}.

\bibitem[Schaeffer et~al.(2023)Schaeffer, Miranda, and Koyejo]{Schaeffer2023AreEA}
Rylan Schaeffer, Brando Miranda, and Oluwasanmi Koyejo.
\newblock Are emergent abilities of large language models a mirage?
\newblock \emph{ArXiv}, abs/2304.15004, 2023.
\newblock URL \url{https://api.semanticscholar.org/CorpusID:258418299}.

\bibitem[Srivastava et~al.(2022)Srivastava, Rastogi, Rao, Shoeb, Abid, Fisch, Brown, Santoro, Gupta, Garriga-Alonso, Kluska, Lewkowycz, Agarwal, Power, Ray, Warstadt, Kocurek, Safaya, Tazarv, Xiang, Parrish, Nie, ..., and Wu]{Srivastava2022BeyondTI}
Aarohi Srivastava, Abhinav Rastogi, Abhishek Rao, Abu Awal~Md Shoeb, Abubakar Abid, Adam Fisch, Adam~R. Brown, Adam Santoro, Aditya Gupta, Adri{\`a} Garriga-Alonso, Agnieszka Kluska, Aitor Lewkowycz, Akshat Agarwal, Alethea Power, Alex Ray, Alex Warstadt, Alexander~W. Kocurek, Ali Safaya, Ali Tazarv, Alice Xiang, Alicia Parrish, Allen Nie, ..., and Ziyi Wu.
\newblock Beyond the imitation game: Quantifying and extrapolating the capabilities of language models.
\newblock \emph{ArXiv}, abs/2206.04615, 2022.
\newblock URL \url{https://api.semanticscholar.org/CorpusID:263625818}.

\bibitem[Vig and Belinkov(2019)]{Vig2019AnalyzingTS}
Jesse Vig and Yonatan Belinkov.
\newblock Analyzing the structure of attention in a transformer language model.
\newblock In \emph{BlackboxNLP@ACL}, 2019.
\newblock URL \url{https://api.semanticscholar.org/CorpusID:184486755}.

\bibitem[Wang et~al.(2022)Wang, Variengien, Conmy, Shlegeris, and Steinhardt]{wang2022interpretability}
Kevin Wang, Alexandre Variengien, Arthur Conmy, Buck Shlegeris, and Jacob Steinhardt.
\newblock Interpretability in the wild: a circuit for indirect object identification in gpt-2 small.
\newblock \emph{arXiv preprint arXiv:2211.00593}, 2022.

\bibitem[Wei et~al.(2022{\natexlab{a}})Wei, Tay, Bommasani, Raffel, Zoph, Borgeaud, Yogatama, Bosma, Zhou, Metzler, hsin Chi, Hashimoto, Vinyals, Liang, Dean, and Fedus]{Wei2022EmergentAO}
Jason Wei, Yi~Tay, Rishi Bommasani, Colin Raffel, Barret Zoph, Sebastian Borgeaud, Dani Yogatama, Maarten Bosma, Denny Zhou, Donald Metzler, Ed~Huai hsin Chi, Tatsunori Hashimoto, Oriol Vinyals, Percy Liang, Jeff Dean, and William Fedus.
\newblock Emergent abilities of large language models.
\newblock \emph{ArXiv}, abs/2206.07682, 2022{\natexlab{a}}.
\newblock URL \url{https://api.semanticscholar.org/CorpusID:249674500}.

\bibitem[Wei et~al.(2022{\natexlab{b}})Wei, Wang, Schuurmans, Bosma, hsin Chi, Xia, Le, and Zhou]{Wei2022ChainOT}
Jason Wei, Xuezhi Wang, Dale Schuurmans, Maarten Bosma, Ed~Huai hsin Chi, F.~Xia, Quoc Le, and Denny Zhou.
\newblock Chain of thought prompting elicits reasoning in large language models.
\newblock \emph{ArXiv}, abs/2201.11903, 2022{\natexlab{b}}.
\newblock URL \url{https://api.semanticscholar.org/CorpusID:246411621}.

\bibitem[Wiegreffe and Pinter(2019)]{Wiegreffe2019AttentionIN}
Sarah Wiegreffe and Yuval Pinter.
\newblock Attention is not not explanation.
\newblock In \emph{Conference on Empirical Methods in Natural Language Processing}, 2019.
\newblock URL \url{https://api.semanticscholar.org/CorpusID:199552244}.

\bibitem[Yuan et~al.(2023)Yuan, Yuan, Li, Dong, Tan, and Zhou]{Yuan2023ScalingRO}
Zheng Yuan, Hongyi Yuan, Cheng Li, Guanting Dong, Chuanqi Tan, and Chang Zhou.
\newblock Scaling relationship on learning mathematical reasoning with large language models.
\newblock \emph{ArXiv}, abs/2308.01825, 2023.
\newblock URL \url{https://api.semanticscholar.org/CorpusID:260438790}.

\bibitem[Zellers et~al.(2018)Zellers, Bisk, Farhadi, and Choi]{Zellers2018FromRT}
Rowan Zellers, Yonatan Bisk, Ali Farhadi, and Yejin Choi.
\newblock From recognition to cognition: Visual commonsense reasoning.
\newblock \emph{2019 IEEE/CVF Conference on Computer Vision and Pattern Recognition (CVPR)}, pages 6713--6724, 2018.
\newblock URL \url{https://api.semanticscholar.org/CorpusID:53734356}.

\end{thebibliography}
\bibliographystyle{plainnat}

\appendix
\newpage
\section{Dataset}
The datasets for syntactical and semantic analysis are generated using templates which are detailed in Table \ref{tab:syntax-data} and in Table \ref{tab:act-patch-data} respectively.

\begin{table*}[h]
\scalebox{0.9}{
\begin{tabular}{clc}
\toprule
\textbf{Template id}&\textbf{Template}&\textbf{Type}\\
\midrule
\multirow{3}{*}{1}&
Alice went to <location> because she wanted to <verb> <object>&B->A\\
&Alice went to <location> and she <verb> <object>&non-causal\\
&Alice went to <random> because she <verb> <object>&random\\
\midrule
\multirow{3}{*}{2}&
Alice went to <location> because <location> is a good place for <object>&B->A\\
&Alice went to <location>  and <location> is <adjective>&non-causal\\
&Alice went to <location> because <location> is a good place for <random>&random\\
\midrule
\multirow{3}{*}{3}&
Alice play <object>  because she enjoys <verb> <object>&B->A\\
&Alice play <object>  and <pronoun> is <adjective>&non-causal\\
&Alice play <object>  because she enjoys <verb> <random>&random\\
\midrule
\multirow{3}{*}{4}&
Bob and Chris made <object> so <pronoun> are <adjective1> and <adjective2>&A->B\\
&Bob and Chris made <object> while <pronoun> are <adjective1> and <adjective2>&non-causal\\
&Bob and Chris made <object> so <pronoun> are <random> and <adjective2>&random\\
\midrule
\multirow{3}{*}{5}&
Bob and Chris got work to do so they are <adjective> to <verb>&A->B\\
&Bob and Chris got work to do but they are <adjective> to <verb>&non-causal\\
&Bob and Chris got work to do so they are <random> to <verb>&random\\
\midrule
\multirow{3}{*}{6}&
Alice went to <location> because she wanted to <verb> <object>&B->A\\
&Alice went to <location> and she <verb> <object>&non-causal\\
&Alice went to <random> because she <verb> <object>&random\\
\midrule
\multirow{3}{*}{7}&
Alice went to <location> because she wanted to <verb> <object>&B->A\\
&Alice went to <location> and she <verb> <object>&non-causal\\
&Alice went to <random> because she <verb> <object>&random\\
\midrule
\multirow{3}{*}{8}&
Alice went to <location> because she wanted to <verb> <object>&B->A\\
&Alice went to <location> and she <verb> <object>&non-causal\\
&Alice went to <random> because she <verb> <object>&random\\
\midrule
\multirow{3}{*}{9}&
Alice went to <location> because she wanted to <verb> <object>&B->A\\
&Alice went to <location> and she <verb> <object>&non-causal\\
&Alice went to <random> because she <verb> <object>&random\\
\midrule
\multirow{3}{*}{10}&
Alice went to <location> because she wanted to <verb> <object>&B->A\\
&Alice went to <location> and she <verb> <object>&non-causal\\
&Alice went to <random> because she <verb> <object>&random\\
\bottomrule
\end{tabular}
}
\caption{Additional Dataset Template for Exploratory Analysis}
\label{tab:syntax-data}
\end{table*}

\begin{table*}[h]

\scalebox{0.9}{
\begin{tabular}{clcc}
\toprule
\textbf{Id}&\textbf{Template}&\textbf{Task Type}&{\textbf{\#}}\\
\midrule
\multirow{1}{*}{ALB}&
John had to [ACTION] because he is going to the [LOCATION].& Action $\xleftarrow{\text{because}}$ Location&6225\\
\midrule
\multirow{1}{*}{AOB}&
Jane will [ACTION] it because John is getting the [OBJECT].& Action $\xleftarrow{\text{because}}$ Object&7509\\
\midrule
\multirow{1}{*}{ALS}&
Mary went to the [LOCATION] so she wants to [ACTION].& Location $\xrightarrow{\text{so}}$ Action&4843\\
\midrule
\multirow{1}{*}{ALS-2}&
Nadia will be at the [LOCATION] so she will [ACTION].& Location $\xrightarrow{\text{so}}$ Action&5600\\
\midrule
\multirow{1}{*}{AOS}&
Sarah wanted to [ACTION] so Mark decided to get the [OBJECT]& Action $\xrightarrow{\text{so}}$ Object&6755\\
\bottomrule
\end{tabular}
}
\caption{Dataset Templates for Causal Relation Prediction}
\label{tab:act-patch-data}
\end{table*}

\section{Proportion of Attention Paid to Delimiters}
Heatmap of the proportion of attention paid to causal delimiters such as "because" and "so" in GPT-2.
\begin{figure}[H]
\centering
\begin{subfigure}{0.3\textwidth}
   \includegraphics[width=1\linewidth]{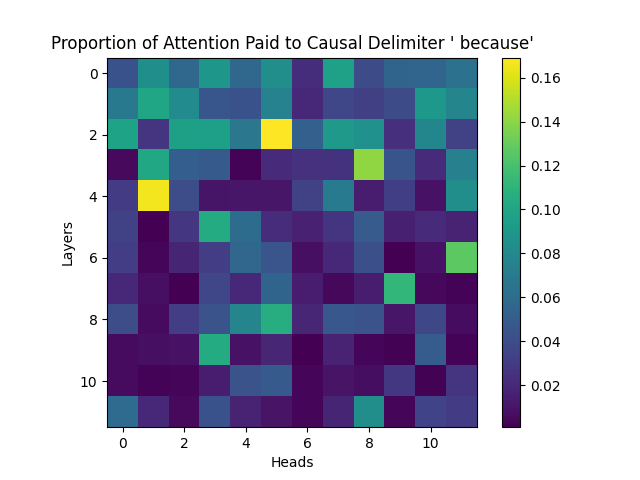}
   \caption{Because}
   \label{fig:att-frac} 
\end{subfigure}
\begin{subfigure}{0.3\textwidth}
   \includegraphics[width=1\linewidth]{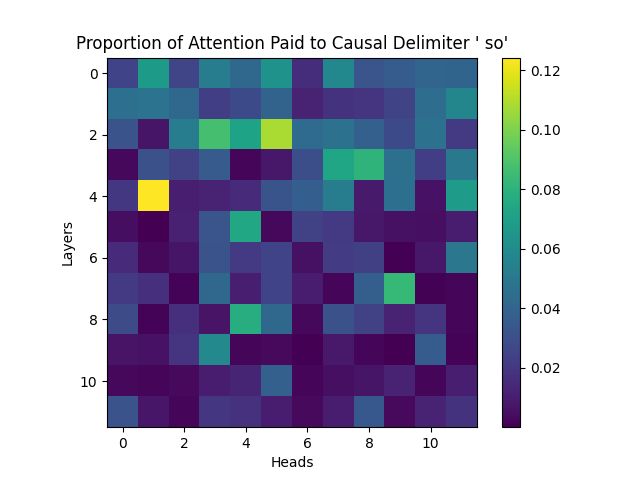}
   \caption{So}
   \label{fig:att-frac} 
\end{subfigure}
\begin{subfigure}{0.3\textwidth}
   \includegraphics[width=1\linewidth]{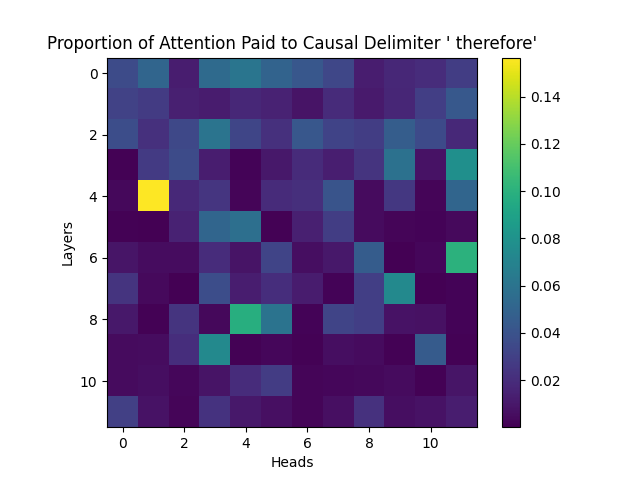}
   \caption{Therefore}
   \label{fig:att-frac} 
\end{subfigure}
\begin{subfigure}{0.3\textwidth}
   \includegraphics[width=1\linewidth]{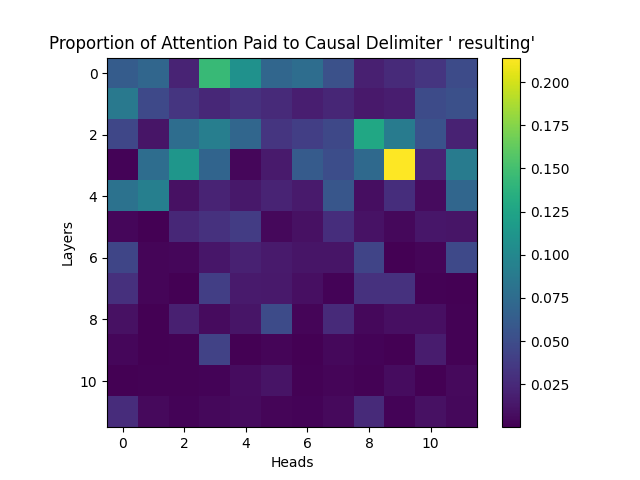}
   \caption{Resulting}
   \label{fig:att-frac} 
\end{subfigure}
\begin{subfigure}{0.3\textwidth}
   \includegraphics[width=1\linewidth]{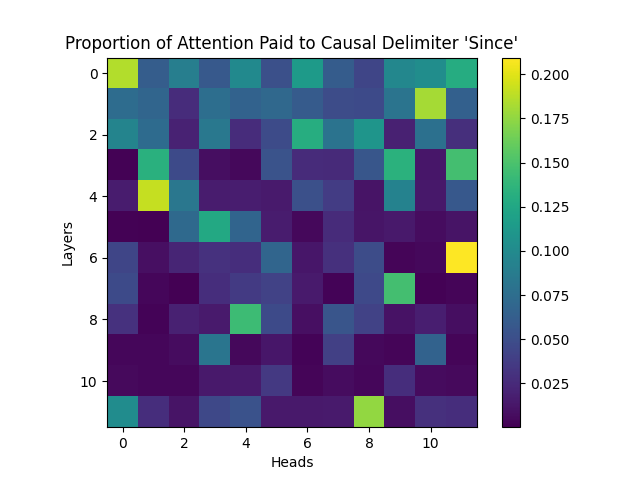}
   \caption{Since}
   \label{fig:att-frac} 
\end{subfigure}
\caption{Proportion of Attention Paid to Causal Delimiters.}
\label{fig:attn-syntax}
\end{figure}

\section{Logit Analysis with Semantic Perturbation}
\begin{figure}[H]
\centering
\begin{subfigure}[b]{0.8\textwidth}
   \includegraphics[width=1\linewidth]{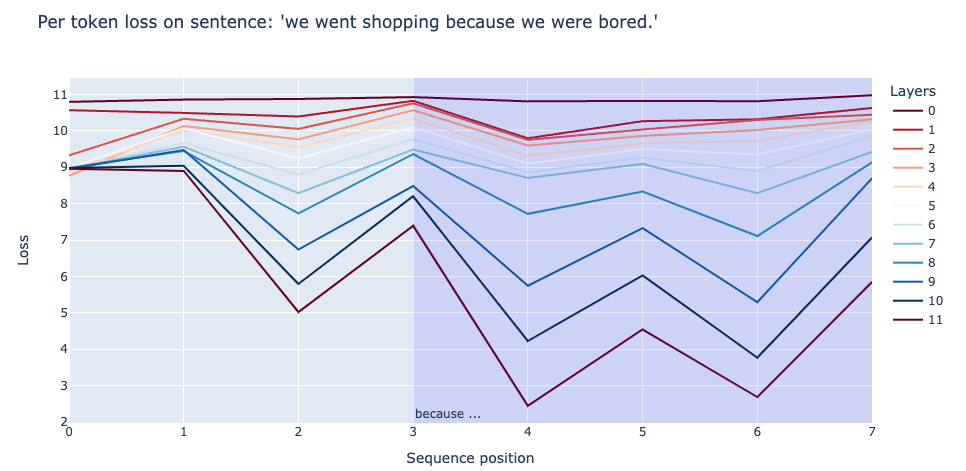}
   \caption{Per layer logit analysis of causal sentence}
   \label{fig:pll-causal} 
\end{subfigure}

\begin{subfigure}[b]{0.8\textwidth}
   \includegraphics[width=1\linewidth]{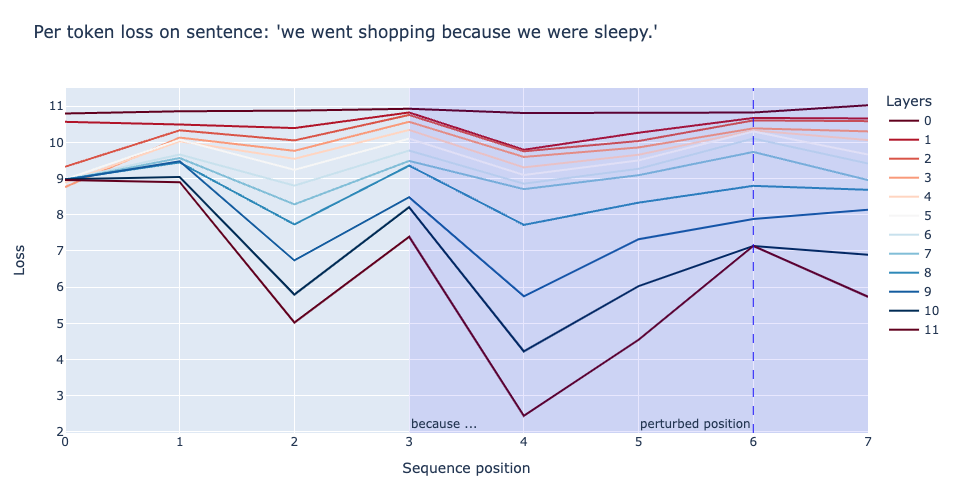}
   \caption{Per layer logit analysis of causally perturbed sentence}
   \label{fig:pll-noncausal}
\end{subfigure}

\caption{Logit analysis with per layer loss.}
\label{fig:pll}
\end{figure}

\section{Activation Patching By Model Components}
\begin{figure}[htbp]
    \centering
    \includegraphics[width=0.9\textwidth]{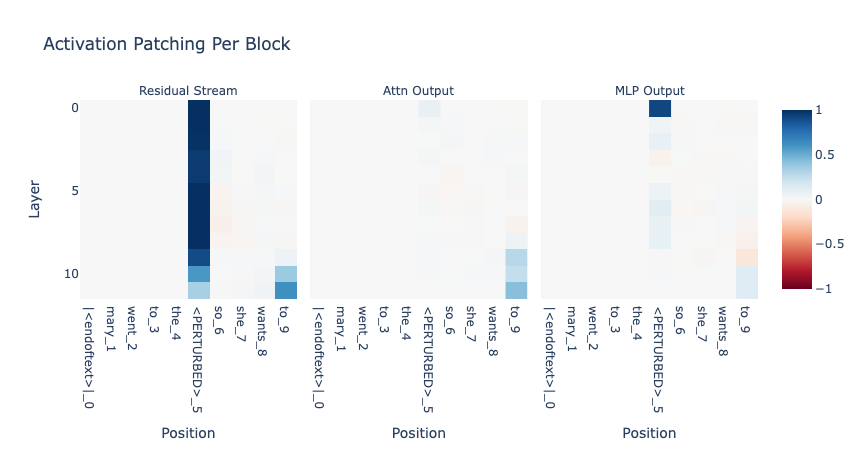}
    \caption{ALS template}
    \label{fig:als_template}
\end{figure}

\begin{figure}[htbp]
    \centering
    \includegraphics[width=0.9\textwidth]{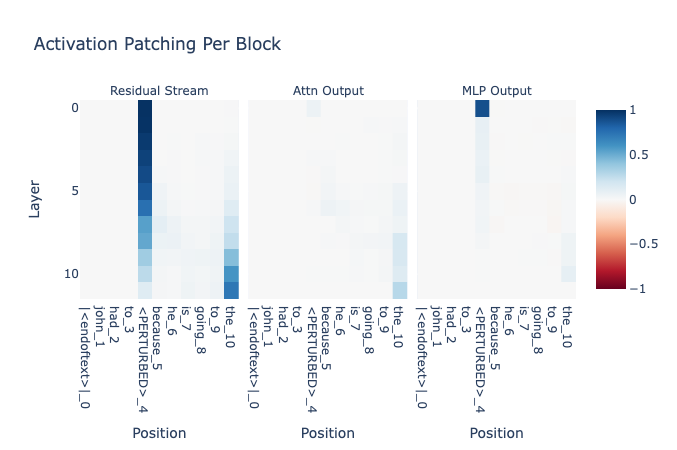}
    \caption{ALB template}
    \label{fig:alb_template}
\end{figure}

\begin{figure}[htbp]
    \centering
    \includegraphics[width=0.9\textwidth]{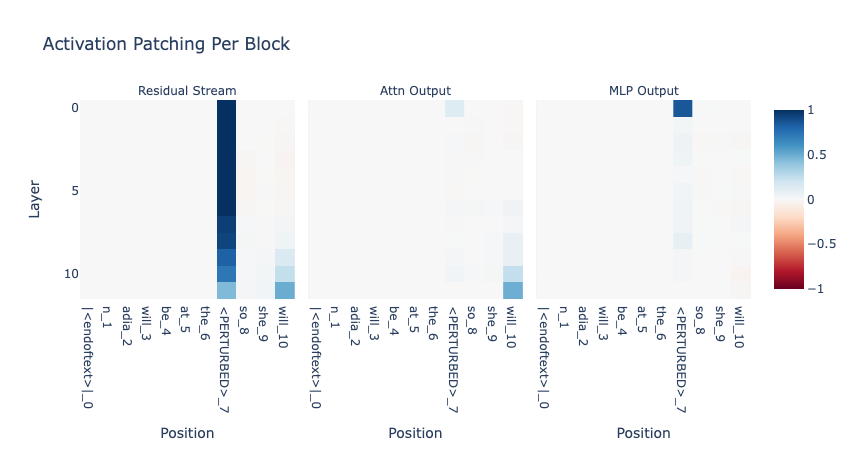}
    \caption{ALS-with template}
    \label{fig:als_with_template}
\end{figure}

\begin{figure}[htbp]
    \centering
    \includegraphics[width=0.9\textwidth]{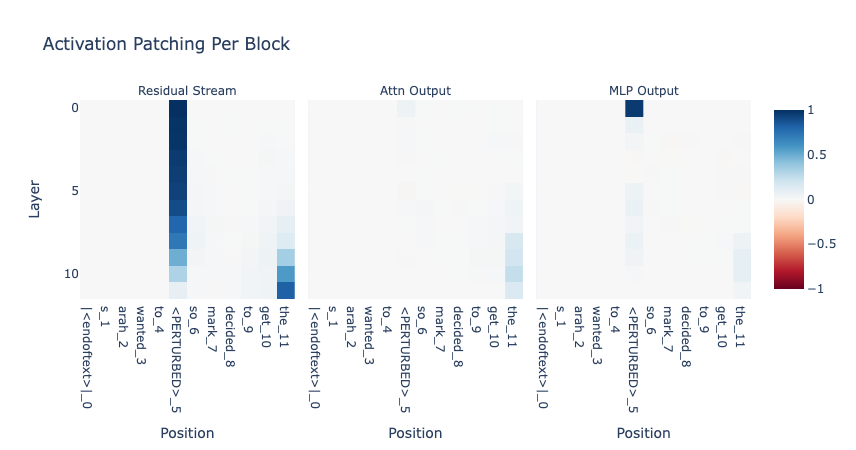}
    \caption{AOS template}
    \label{fig:aos_template}
\end{figure}

\begin{figure}[htbp]
    \centering
    \includegraphics[width=0.9\textwidth]{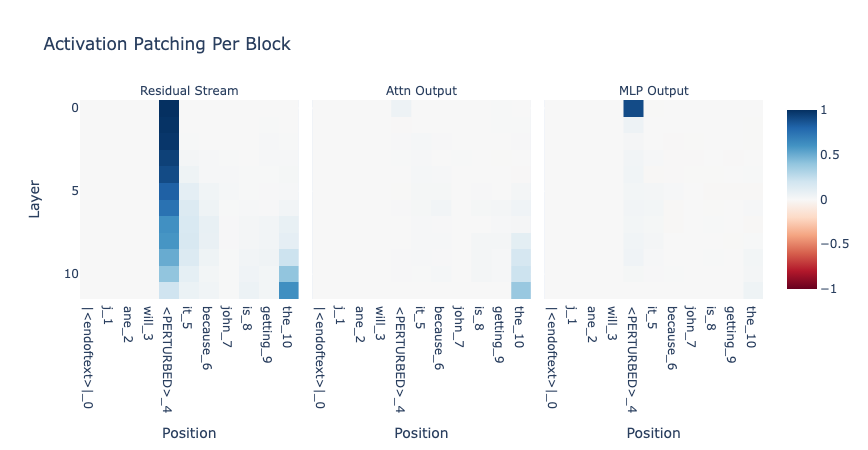}
    \caption{AOB template}
    \label{fig:aob_template}
\end{figure}

%%%%%%%%%%%%%%%%%%%%%%%%%%%%%%%%%%%%%%%%%%%%%%%%%%%%%%%%%%%%

\end{document}